\documentclass{article}
\usepackage{ijcai19}

\usepackage{times}
\usepackage{soul}
\usepackage{url}
\usepackage[utf8]{inputenc}
\usepackage[small]{caption}
\usepackage{graphicx}
\usepackage{amsmath}
\usepackage{booktabs}
\usepackage{multirow}
\usepackage{algorithm}
\usepackage{algorithmic}
\usepackage{amsfonts}
\usepackage{color}

\title{Semi-supervised Learning with Contrastive Predicative Coding}

\author{
Jiaxing Wang$^1$
\and
Yin Zheng$^2$\and
Xiaoshuang Chen$^{3}$\and
Junzhou Huang$^2$\And
Jian Cheng$^1$
\affiliations
$^1$Institute of
Automation, Chinese Academy of Sciences\\
$^2$ Tencent\\
$^3$ Tsinghua University\\
}

\begin{document}

\maketitle

\begin{abstract}
Semi-supervised learning (SSL) provides a powerful framework for leveraging unlabeled data when labels are limited or expensive to obtain. SSL algorithms based on deep neural networks have recently proven successful on standard benchmark tasks. However, many of them have thus far been either inflexible, inefficient or non-scalable. This paper explores recently developed contrastive predictive coding technique to improve discriminative power of deep learning models when large portion of labels are absent. Two models, cpc-SSL and a class conditional variant~(ccpc-SSL) are presented. They effectively 
exploit the unlabeled data by extracting shared information between different parts of the (high-dimensional) data. The proposed approaches are inductive, and scale well to very large datasets like ImageNet, making them good candidates in real world large scale applications.



\end{abstract}

\section{Introduction}
Semi-supervised learning (SSL) is drawing great attention due to the increasing size of modern datasets and the high cost of obtaining label information. Let $D_L$  be a set of labeled data points, and $D_U$ be a set of unlabeled data. By utilizing both $D_L$ and $D_U$ ($|D_L| \ll |D_U|$),  SSL effectively learns model that generalizes better than models learned from labeled data $D_L$ only

Various approaches to SSL have been proposed (See section \ref{sec:related works} for an overview). They can be roughly categorized into graph-based methods, perturbation-based methods and generative model based methods. Graph-based methods operate over a graph 
where data points are represented as vertices, and edges encode the similarity between the labeled and unlabeled instances. \cite{zhu02lp} proposed label propagation, which iteratively propagates the class posterior of a node to its neighbours until the process finally reach an equilibrium. Many variations followed~\cite{haeusser17ssl,konstantinos18ssl}, yet the super-quadratic computational complexity hinders their use in modern large scale applications. Perturbation based methods tries to regularize the input-output mapping to be consistent when noise is applied to the input, which has been shown effective in SSL~\cite{bachman14pseudo-ensumble}. These methods scale well and demonstrate potential on large scale benchmarks like ImageNet~\cite{tarvainen17meanteacher}. However, like self-supervision~\cite{scudder65self-sup}, perturbation based methods use the model generated labels to guide the training process, thus induce confirmation bias. Generative models acquire the latent manifold’s structure of data by modelling the joint distribution $p(x, y) = p(y | x)p(x)$. Estimating $p(x)$, however, makes the solution sub-optimal for SSL as it unnecessarily spends representational power preserving (to preserve) the very details of the inputs, e.g. pixel level information in images, which also harms the scalability of the model. Models that utilize more globle structures of data rather than the local features is urgently needed. Although previous works has shown promising results on several benchmarks and in certain domains, SSL method that is practical to real world applications is still left largely unexplored.
 
Predictive coding uses contextual data at hand to predict future or missing information. With this idea, ~\cite{mikolov13word2vec} learns distributed word representations by predicting neighboring words. More recently,~\cite{aaron18cpc} use a few patches of data to predict the other patches for general unsupervised representation learning. The proposed method is named "constrastive predictive coding" (CPC), which is able to extract meaningful representations in various domains, including speech, images and text. Success of these methods lie in that the contexts are often conditionally dependent on the same shared high-level latent information with the predicted future values. By solving this prediction problem, we automatically infer these salient latent factors. CPC discards very local information, only concentrates on more global structures of data, which makes CPC an ideal candidate for developing SSL models.

In this paper, we propose semi-supervised contrastive predictive coding (cpc-SSL). We infer the salient latent factors that encode the underlying shared information between different parts of the (high-dimensional) data, and regularize them to improve label prediction. We further propose a class-conditional variant of cpc-SSL (ccpc-SSL). Compared with cpc-SSL, ccpc-SSL disentangles the label information from the latent variables and separates the pathways of inferring labels and context variables. Empirical results on widely different data modalities demonstrate that: (1) the proposed models can significantly improve the prediction performance when only limited labeled data is available. (2) The proposed models scale well to very large datasets, which poses great challenge to many previous SSL methods.

The rest of the paper is organized as follows. We briefly review contrastive predictive coding and introduce our proposed cpc-SSL and ccpc-SSL in Section \ref{sec:method}. In Section \ref{sec:experiments}, we demonstrate the experiment results and compare our method with state of the arts on both image recognition and sentiment classification tasks. Finally, we review related works in Section \ref{sec:related works}. 

\section{Method}
\label{sec:method}
\subsection{Contrastive Predictive Coding}
Contrastive predictive coding(CPC) is recently introduced for unsupervised representation learning. It learns the salient latent factors that encode the underlying shared information between different parts of the (high-dimensional) signal. Advantage of contrastive predictive coding over generative models, which has been extensively exploited in label absent cases,  lies in that CPC discards low-level local information and noise that is unrelevent to down streaming tasks. 
\begin{figure}[h]
\hfill
\begin{center}
\includegraphics[scale=0.7]{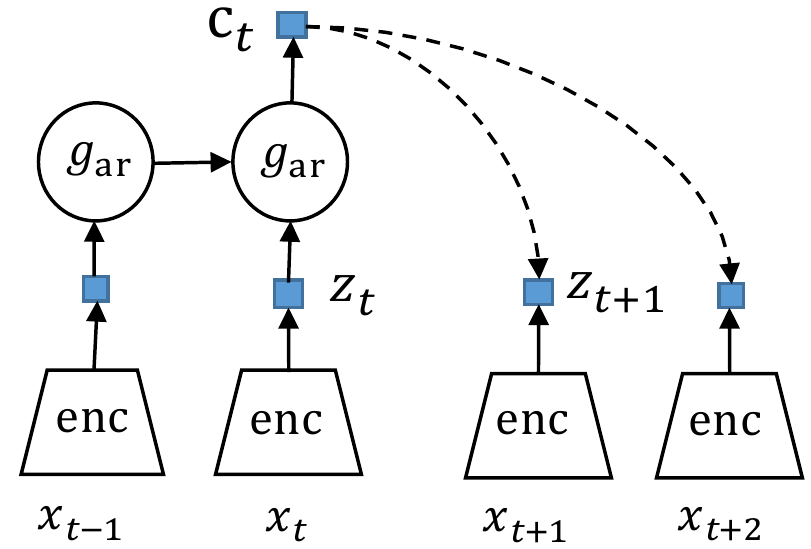}
\end{center}
\caption{Overview of Contrastive Predictive Coding, }
\label{fig:cpc-diag}
\end{figure}
Figure \ref{fig:cpc-diag} shows the general architecture of contrastive predictivec coding models. In ~\cite{aaron18cpc}, a non-linear encoder is used to map the sequence of observations $x$ to a sequence of latent representations $z = enc(x)$. Next, an autoregressive model $g_{ar}$ summarizes all $z_{\leq t}$ in the latent space and produces a latent representation $c_t = g_{ar}(z_{\leq t})$. 

Let us assume altogether we predict up to $K$ time-steps with the context representation $c_t$  . For each prediction time step k, a set $X$ is constructed for noise contrastive estimation. $X$ contains one positive sample $\hat{x}$ from $p(x_{t+k}|c_t)$ and $N - 1$ negative samples from the proposal distribution $p(x_{t+k})$. The goal of noise contrastive estimation is to correctly pick the positive sample which comes from $p(x_{t+k}|c_t)$. The noise estimation objective is:
\begin{equation}
    L_N = -\mathbb{E}_{X} \left[\log \frac{f_k(x_{t+k}, c_t)}{\sum_{x_j \in X} f_k(x_j, c_t)} \right] \label{eq:nce_loss}
\end{equation}

where $f_k(x_{t+k}, c_t)$ gives the normalized probability of recognizing $x_{t+k}$ to be the positive sample. $f_k(x_{t+k}, c_t)$ can be simply a bilinear model:
\begin{equation}
    f_k(x_{t+k}, c_t) = \exp({z_{t+k}^T W_k c_t})
\end{equation}


As shown in~\cite{aaron18cpc}, optimizing~(\ref{eq:nce_loss}) maximizes a lower bound of mutual information between $c_t$ and $z_{t+k}$, thus the model is forced to capture the underlying shared information between different time-steps. Theoretically validates the exploitation of contrastive predicative coding for representation learning of the whole sequence.  

\subsection{Contrastive Predicitive Coding for Semi-Supervised Learning}
We extend the contrastive predicitive coding method for semi-supervised learning. Two models, semi-supervised learning with  contrastive predicitive coding(cpc-SSL) and its class-conditional variant ccpc-SSL, are proposed. For both models, we present efficient learning and inference algorithms.

\begin{figure}[h]
\hfill
\begin{center}
\includegraphics[scale=0.6]{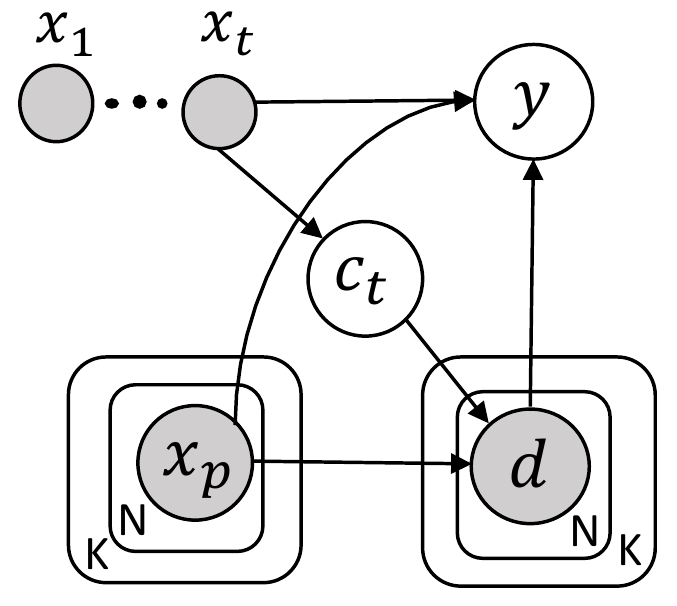}
\end{center}
\caption{Probabilistic model for cpc-SSL}
\label{fig:cpc-semi-cpc-deterministic}
\end{figure}

Without loss of generality, we consider the multiclass classification\footnote{The model can be easily adopted to other tasks, e.g. image captioning, structured prediction.}, where $y \in C =\{1, . . . , M\}$.  A cpc-SSL consists of two components: (1) a contrastive predictive coding model to describe input data and infer salient latent factors; and (2) a multi-class classifier for classification. The first component helps explore the intrinsic structures and provides a latent representation of the data. The latent representations allow for a clustering of related observations in the latent space, which facilitates separation, even with a limited number of labels. The proposed cpc-SSL is illustrated in Fig \ref{fig:cpc-semi-cpc-deterministic}. $x$ denotes the observed contextual data. $x_p$ is the future sample to be predicted within set $X$ and $d$  a binary variable indicating whether $x_p$ is the positive sample from $p(x_{t+k}|c_t)$. cpc-SSL use aggregation of both the contextual data $x$ and $x_p$  to predict the label $y$ of the whole sequence.  

The probabilistic model we use is:
\begin{align}
\begin{split}
	& p(c_t | x_{\leq t}) =  \int p(c_t | z_{\leq t} ) p(z_{\leq t} | x_{\leq t}) dz_{\leq t} \\
	& p(d | c_t, x_p) = Ber(f(d ; c_t, x_p, W)) \\
	& p(y|\widehat{z}_{1:t+K}) =Cat(h(y;\hat{z}_{1:t+K}, \theta))
	\label{eq:cpc-SSL GM}
\end{split}
\end{align}
where Cat(·) is a multinomial distribution and Ber(·) denotes Bernoulli distribution. $f$ gives unnormalized probability of $x_p$ being the positive sample in current prediction time-step and $h$ is the classifier on the whole sequence. In equation (\ref{eq:cpc-SSL GM}), $p(c_t | x_{\leq t})$ is generally intractable, for simplicity, we just use an empirical distribution $p(z_{\leq t} | x_{\leq t}) = \prod^t \delta_{g(x) - z} $  where $g$ is the nonlinear encoder. The latent variable $z$ in the graphical model Figure~\ref{fig:cpc-semi-cpc-deterministic} is thus omitted. To correctly classify the sequence, only  the true underlying data representation is need, the noise contrastive samples will not be evolved in the classifier. We thus build the classifier $p(y|z_{\leq t}, z_p) = p(y|\widehat{z}_{1:t+K})$, where $\hat{z}$ denotes the latent representation of the positive sample in noise contrastive prediction. 

For labeled data, likelihood for a single data sequence is:
\begin{align}
\begin{split}
	\mathcal{L} & =  \log p(y |d, x_{\leq t}, X) + \log p(d | x_{\leq t}, X) \\
	                     & = \log p(y | \hat{z}_{1:t+K}) +  \log \mathbb{E}_{p(c_t | x_{\leq t})} \left[ p(d | c_t, x_p) \right]
    \label{eq:cpc-SSL-loss-l}
\end{split}
\end{align}
This loss consists of two parts, a noisy contrastive estimation loss $ p(d_p | x_{\leq t}, X) $ and a classification loss $ p(y | x_{\leq t}, X) $.  For unlabeled data, we simply ignore the classification loss term.
\begin{align}
	 \mathcal{U} = \log~p(d | x_{\leq t}, X) = \log \mathbb{E}_{p(c_t | x_{\leq t})} \left[ p(d | c_t, x_p) \right]
	 \label{eq:cpc-SSL-loss-u}
\end{align} 
Following~\cite{kingma14dgm,maaloe16adgm}, classification loss is given higher weights so we introduce an explicit classification loss for labeled data.
\begin{equation}
	\mathcal{L}_{cls} =  \mathbb{E}_{\vec{x_l},\vec{y_l}} \left[\log p(y | d, x_{\leq t}, X) \right]
	\label{eq:cls_loss}
\end{equation}
where $\vec{x_l}, \vec{y_l}$ denotes the labeled data(the sequence as a whole). Similarly we use $\vec{x_u}$ to denote the unlabeled data sequence.
The final objective for SSL is then:
\begin{equation}
	\mathcal{J} = \sum_{(\vec{x_l}, \vec{y_l})} \mathcal{L}(\vec{x_l}, \vec{y_l}) + \sum_{\vec{x_u}} \mathcal{U}(\vec{x_u}) + \alpha \mathcal{L}_{cls}(\vec{x_l}, \vec{y_l})
	\label{eq:cpc-SSL}
\end{equation}
Note that evaluating equation (\ref{eq:cpc-SSL-loss-l}) and (\ref{eq:cpc-SSL-loss-u}) asks for computing expectation over aggregated contextual latent factor $c_t$, which is intractable in general. Here we utilize the reparametrization trick proposed  in~\cite{kingma14vae} for efficient inference.

\subsection{Class Conditional Variant of cpc-SSL}
Optimizing (\ref{eq:cpc-SSL}) leads to a weak coupling between the cpc model and the classifier. Though this kind of weakly coupled approach has been shown effective in previous works~\cite{li15mmdgm}, strongly coupled model usually yields better performance in real world application~\cite{kingma14dgm,li15mmdgm}. Below, we present a conditional variant of cpc-SSL (ccpc-SSL), to strongly couple the classifier and contrastive prediction model.

\begin{figure}[h]
\hfill
\begin{center}
\includegraphics[scale=0.6]{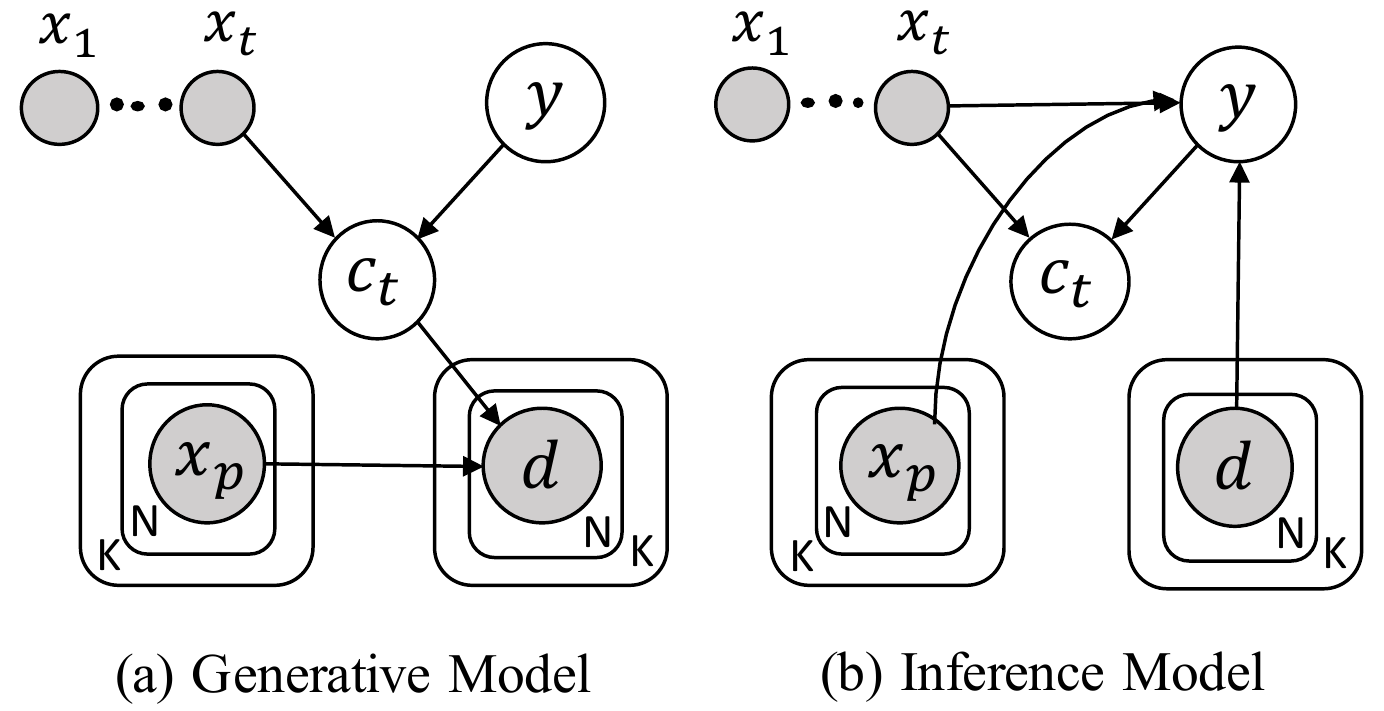}
\end{center}
\caption{Probabilistic graphical model of ccpc-SSL for semi-supervised learning. }
\label{fig:cpc-SSL-graphical}
\end{figure}

As in cpc-SSL, an ccpc-SSL consists of two components: (1) a multi-class classifier to infer labels given inputs and (2) a class-conditional  contrastive predictive coding model to extract salient latent factors. This time, the aggregated contextual factor $c_t$ is generated with the class label $y$ known. The SSL problem is then recognized as a specialised missing data imputation task. Fig.~\ref{fig:cpc-SSL-graphical} describes the graphical models of ccpc-SSL. The incoming joint connections to each variable are complete conditionals modeled with neural networks. Following tradition, networks forming the generative model are parametrized by $\theta$, while parameters in inference model are denoted by $\phi$. Below, we present the learning objective of ccpc-SSL formally, which consists of several key components.

The probabilistic model for ccpc-SSL is :
\begin{align}
\begin{split}
	& p(y)  = Cat(\pi) \\
	& p(c_t | y, x_{\leq t}) = \mathcal{N}(\mu_{\theta}(y, x_{\leq t}), \sigma_{\theta}(y, x_{\leq t})) \\
	& p(d | c_t, x_p) = Ber(f(d; x_p, c_t, W))  \\
\end{split}
\end{align}
y is treated as a latent variable for the unlabeled data points. The inference model is then defined as $ q(c_t, y | d, x_{\leq t}, X) = q(c_t | x_{\leq t}, y) q(y | d, x_{\leq t}, X)$, where:
\begin{align}
\begin{split}
	 & q(y | d, x_{\leq t}, X) = Cat(f(y; \hat{z}_{1:t+K}, \phi)) \\
	 & q(c_t | x_{\leq t}, y) = \mathcal{N}(\mu_{\phi}(y, \hat{z}_{\leq t}), \sigma_{\phi}(y, \hat{z}_{\leq t}))        
\end{split}
\end{align}

We optimize the model by maximizing the lower bound of the likelihood. The variational lower bound on the marginal likelihood for a labeled data sequence is:
\begin{align}
\begin{split}
	 \mathcal{L} = \mathbb{E}_{q(c_t | x_{\leq t}, y)} \left[ \log p(d, c_t | y, x_{\leq t}, X) \right] \\ +\mathcal{H}(q(c_t | x_{\leq t}, y))
\end{split}
\end{align}
where $\mathcal{H}(q(c_t | x_{\leq t}, y))$ is the entropy of approximated posterior $q(c_t | x_{\leq t}, y)$

For the case where the label is missing, the label is treated as a latent variable over which we perform posterior inference and the resulting bound is:
\begin{align}
\begin{split}
	\mathcal{U} = \mathbb{E}_{q(c_t, y | d, x_{\leq t}, X)} \left[ \log~p(d, c_t, y | x_{\leq t}, X) \right]  \\ + \mathcal{H}(q(c_t, y | d, x_{\leq t}, X))
	\label{eq:ccpc-SSL-unlabeled}
\end{split}
\end{align}

Again, an explicit classification loss, equation (\ref{eq:cls_loss}) for labeled data is applied.
The extended objective function is finally:
\begin{equation}
	\mathcal{J} = \sum_{(\vec{x_l}, \vec{y_l})} \mathcal{L}(\vec{x_l}, \vec{y_l}) + \sum_{\vec{x_u}} \mathcal{U}(\vec{x_u}) + \alpha \mathcal{L}_{cls}(\vec{x_l}, \vec{y_l})
\end{equation}

Note that evaluation of equation (\ref{eq:ccpc-SSL-unlabeled}) asks for expectation over class label y. Making it scales linearly in the number of classes. Re-evaluating the likelihood for each class during training is prohibitively expensive when the number of classes is large. Reparameterization trick is not directly applicable on discrete latent variables. Here we adopt Gumble-Softmax trick~\cite{eric15gumble}. Gumbel Softmax distribution is a continuous relaxation of categorical distribution controlled by a temperature parameter. During training, samples y from this relaxed distribution is used to evaluate $\ref{eq:ccpc-SSL-unlabeled}$. As the temperature anneal, this distribution asymptotically converges to the true underlying categorical distribution $q(y | d, x_{\leq t}, X)$. Gumbel Softmax trick allows gradients flow trough the random variable $y$.

\subsection{Computational Complexity}
The computational cost of cpc-SSL is slightly larger than that of a 'standard' deep neural network classifier. The exact algorithmic complexity depends on the way the data sequence is constructed. Here we give a sketchy computational complexity analysis. 

During training, the overall algorithmic complexity of a single update of the parameters is $C_{tr} = M(NK+t)C_{enc} + MC_{ag} +MC_{cls}$ while that of a 'standard' classifier is $M(K+t)C_{enc} + MC_{cls}$. $M$ is the minibatch size. $N$ is the number of data in contrastive set $X$, $K$ represents the number of time-steps to be predicted and $t$ is the number of contextual data in the sequence. $C_{enc}$ is the computational cost of encoding data patch $x$. The encoder is typically a  deep neural network, which forms the computational bottleneck. $C_{cls}$ is the classification layer cost. cpc-SSL contains a auto-regressive model to compute the aggregated contextual representation $c_t$, which introduces $C_{ag}$ additional complexity. However, $C_{ag}$ is usually much smaller compared to $C_{enc}$, as the auto-regressive model operates in the latent space instead of high dimensional data space. For testing, the algorithmic complexity of a feed forward procedure is $C_{enc} + C_{ag} +C_{cls}$, which is nearly the same as a 'standard' deep neural network classifier. The class-conditional variant ccpc-SSL shares the same algorithmic complexity with cpc-SSL if Gumbel-Softmax trick is adopted. These complexities make the proposed approaches appealing, since they are no more expensive than a 'standard' deep neural network classifier in inference time.

\section{Experiments}
\label{sec:experiments}
We now present the semi-supervised classification results on two different domains: image recognition and sentiment classification. 

\subsection{Image Benchmarks}
\begin{figure}[h]
\hfill
\begin{center}
\includegraphics[scale=0.4]{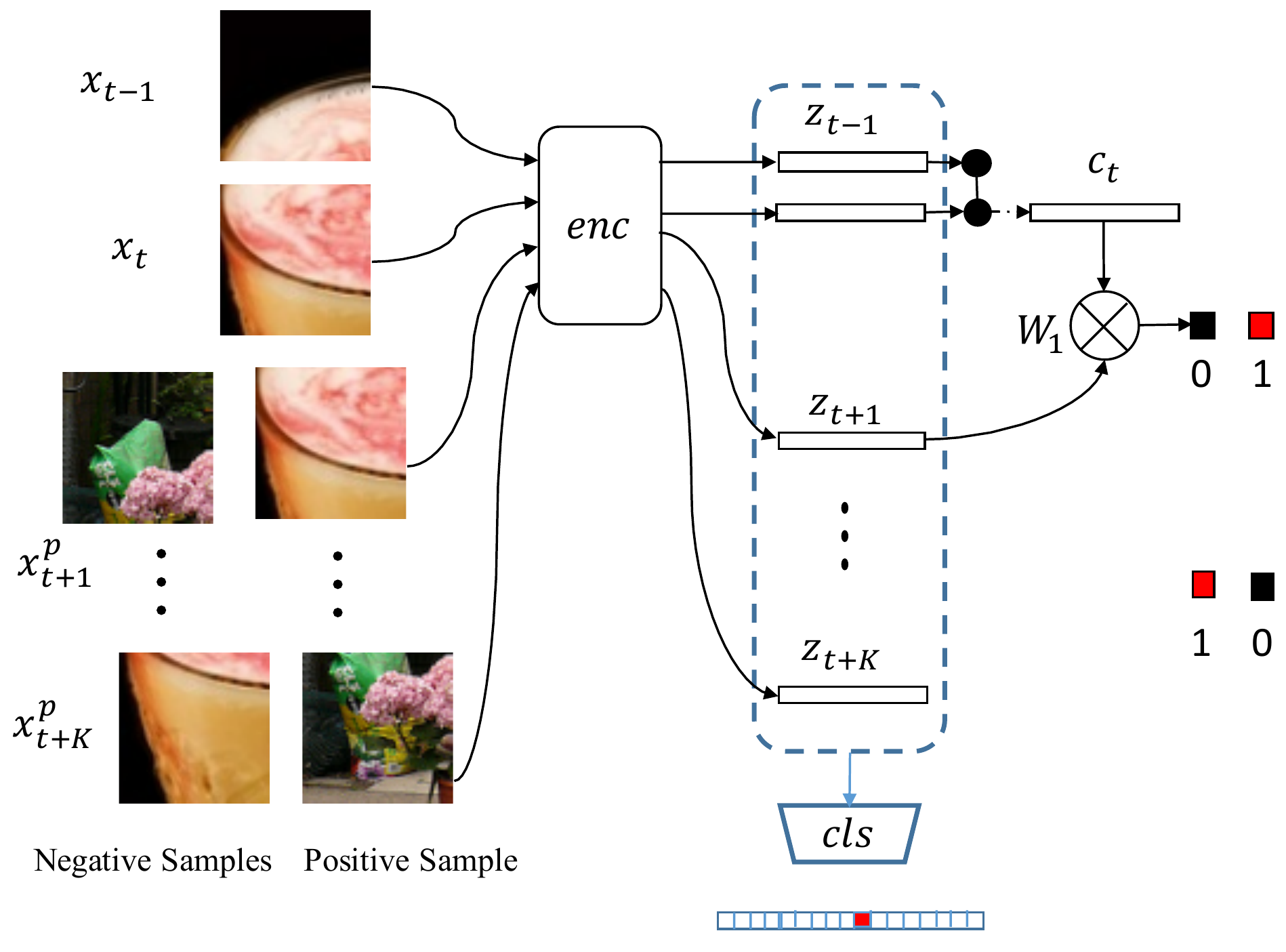}
\end{center}
\caption{Overview of cpc-SSL for image recognition. Blue lines and dashed box are for classification and black  solid arrows are for contrastive predictive coding}
\label{fig:cpc-SSL-img}
\end{figure}

\begin{table*}[]
\resizebox{\textwidth}{14mm}{
\begin{tabular}{@{}lccccccccr@{}}
\specialrule{0em}{0.5pt}{1pt}
\toprule
\multirow{2}{*}{} & \multicolumn{4}{c}{Top 1 Acc.} & \multicolumn{4}{c}{Top 5 Acc.} & Test Time \\ \cmidrule(lr){2-9}
                                                  & 1 \%  & 5 \%  & 10 \%  & 20 \% & 1 \%  & 5 \%  & 10 \%  & 20 \% &           \\ \midrule
BSVM~\cite{pu16semi-vae}      & 43.98$\pm$1.15 & 47.36$\pm$0.91 & 48.41$\pm$0.76 & 51.51$\pm$0.28 & 60.57$\pm$1.61 & 62.67$\pm$1.14 & 64.76$\pm$0.90 & 75.67$\pm$0.19 & -        \\
Mean Teacher~\cite{tarvainen17meanteacher}  & 52.27$\pm$0.89 & 59.47$\pm$0.57 & 63.79$\pm$0.36 & \textbf{66.30$\pm$0.31} & \textbf{78.51$\pm$0.94} & 82.53$\pm$0.64 & 84.85$\pm$0.42       & \textbf{88.02$\pm$0.31}      & 10.48   \\
Supervised-Only                     & 39.64$\pm$0.73 & 51.21$\pm$0.52 & 57.25$\pm$0.51 & 61.24$\pm$0.26 & 57.45$\pm$1.02 &      73.84$\pm$0.68 &  77.87$\pm$0.69      & 81.98$\pm$0.21      & 10.23  \\
Cpc-SSL(Ours)                       & \textbf{54.53$\pm$1.06} & 60.17$\pm$1.21      & \textbf{63.86$\pm$0.72} & 66.13$\pm$0.41 & 78.38$\pm$1.24 & 82.44$\pm$0.96    & \textbf{85.03$\pm$0.78}    & 87.89$\pm$0.37      &  19.56       \\
Ccpc-SSL(Ours)                     & 54.21$\pm$0.92 &  \textbf{60.24$\pm$0.90}     & 63.67$\pm$0.75 & 66.15$\pm$0.37      & 78.32$\pm$1.07      & \textbf{82.78$\pm$1.01}      & 84.79$\pm$0.81       &  87.64$\pm$0.41  & 19.56          \\ \bottomrule
\end{tabular}}
\caption{Semi-supervised classification accuracy (\%) and testing time (ms per image) on the validation set of ImageNet 2012. The results for BSVM is taken from~\protect\cite{pu16semi-vae}, which may not be directly comparable because of architectural differences. BSVM maintains both a convolutional encoder and a decoder, which may cause great computational overhead. For mean teacher, we ran the code given in~\protect\cite{tarvainen17meanteacher}, the model architecture and training protocol is the same with ours. }
\label{tab:cpc-SSL-ImageNet}
\end{table*}

\begin{table}
\centering
\begin{tabular}{lrrrr}  
\toprule
Scenario  & 1 \% & 5 \% & 10 \% & 20 \% \\
\midrule
Supervised-only  & 61.24  & 68.47 & 73.85  & 78.18    \\
Cpc-SSL    & 66.57  & \textbf{74.86} & 77.34  &  \textbf{80.83}    \\
Ccpc-SSL   & \textbf{67.14}  & 74.83  & \textbf{77.51} & 80.81   \\
\bottomrule
\end{tabular}
\caption{Semi-supervised classification accuracy (\%) on Large Movie Review dataset.}
\label{tab:cpc-SSL-LargeMovieReview}
\end{table}
Various SSL methods have been proposed and tested on small and middle sized bench marks like MNIST,  SVHN and CIFAR-10. Many of them, however, are not suitable for large scale real world applications. In our image recognition experiment we use the ILSVRC ImageNet competition dataset~\cite{russakovsky15imagenet}. The ImageNet dataset has been used as large scale benchmark to evaluate semi-supervised vision models in previous works~\cite{pu16semi-vae,tarvainen17meanteacher}. Here, the original images are reshaped and cropped to 256 $\times$ 256. To isolate the effect of the SSL method, no additional data augmentation is applied. 

cpc-SSL on image recognation task is overviewed in Figure \ref{fig:cpc-SSL-img}. Following \cite{aaron18cpc}, From each 256 $\times$ 256 image, we extract a $7 \times 7$ grid of 64 $\times$ 64 crops with 32 pixels overlap. Image crops from a column form a data sequence. Each crop is encoded by the encoder $g_{enc}$ (mapping from x to z) to get a feature vector. We use the first three blocks of ResNet101-v2. Note that this encoder is not pretrained. Batch Norm is also dropped as in \cite{aaron18cpc}. We chose $t=2$ and predict the following 5 patches in a column from top to bottom. $p(c_t|z_{\leq t})$ is modeled with a GRU RNN~\cite{cho14gru} with cell dimension 256. Mean and variance are computed from the final hidden states of the GRU. For classification, The $7 \times 7$ feature vectors $\hat{z}$ from an image are spatially mean-pooled to an aggregated image level feature, which is used as the final input feature for a linear classifier. ccpc-SSL contains additional components $q(y|d,x_{\leq t},X)$ and $q(c_t|x_{\leq t}, y)$. Similar to $p(c_t|z_{\leq t})$, $q(c_t|x_{\leq t}, y)$ is also modeled with a GRU RNN, but with the inferred image label $y$ fed in. For simplicity, the approximated label posterior $q(y|d,x_{\leq t},X)$ is also modeled by a linear classifier on the aggregated image level features. 

Adam optimizer with learning rate 1.6e-4 is used and the model is trained on 8 GPUs. We use batch size 16 on each GPU, within which 8 samples are labeled the other 8 are unlabled. We split the 1.3M training images into a labeled and unlabeled set, and vary the proportion of labeled images from 1 \% to 20 \%. 

Table \ref{tab:cpc-SSL-ImageNet} shows the top-1 and top-5 classification accuracies compared with a supervised-only baseline and other state-of-the-art methods. The supervised-only baseline just fits the labeled data, which is used as a reference to illustrate the benefits of leveraging unlabeled images through contrastive predictive coding. The supervised-only baseline is a conventional convolutional neural network, whose feature extractor is the same with the encoder of cpc-SSL. To avoid serious over-fitting in data scarce case, we apply weight decay with 1e-4 rate and data augmentation including random flipping and cropping (Following the convention training ResNet models).  The supervised baseline shares the same training protocol as cpc-SSL, all the details are kept the same.  BSVM~\cite{pu16semi-vae} is a generative model based SSL method, which use a deep convolutional variational autoencoder to model joint distribution $p(x,y)$. Mean Teacher~\cite{tarvainen17meanteacher} is a perturbation based approach, which is the current state of the art. As shown in Table~\ref{tab:cpc-SSL-ImageNet}, our semi-supervised learning approaches significantly outperform their supervised-only counterpart, and are at least comparable to the state-of-the-art method. For the consideration of inference time, our image crops have 32 pixels overlap, introducing nearly additional $1 \times$ convolution operations compared to the supervised baseline. The problem can be largely alleviated by designing better sequence construction schemes. 

As a reference for training computational cost, our model takes about 5 days to train on ImageNet2012, with 8 P40 GPUs. For real-world application consideration~\cite{oliver18realisticeval},
we didn't do class balance when construct the labeled training set, dataset specific hyper-parameter tuning was neither conducted. Actually the proposed cpc-SSL and ccpc-SSL are not sensitive to the setting of hyperparameters. $\alpha$ is set to be $8\rho$ where $\rho = \frac{|D_U|}{|D_L|}$ for all the experiments. Doing more careful training data construction or introducing delicate hyper-parameter tuning will further boost the performance of our models. 

\subsection{Natural Language Benchmarks}
A sentiment classification experiment is conducted to validate the applicability of the proposed model on neutral languages data. We use Large Movie Review Dataset~\cite{maas11lmr}, which contains 25,000 training movie reviews and 25,000 testing reviews. We split the  training reviews into a labeled and an unlabeled set. The same as before, proportion of labeled data varies from 1 \% to 20 \%.

Our model consists of a simple sentence encoder $g_enc$ (a 1D-convolution + ReLU + max-pooling) that embeds a whole sentence into a vector z. Three groups of filters with size [3, 4, 5] operates on the sentence, resulting in three 128 dimensional features, the 3 features are concatenated to get a sentence representation. Again, we use a GRU to predict up to 6 future sentences from 2 contextual sentences in contrastive noisey estimation. As in image classification task, the final document level representation obtained by mean-pooling sentence features. The classifier model is simply a logistic regression model. We used Adam optimizer with a learning rate of 1e-5 with batch size 64.  Hyper-parameters are set in the same way as in image experiment. 

Table \ref{tab:cpc-SSL-LargeMovieReview} reports the sentiment classification accuracies. Due to lack of comparative results exist on SSL tasks, we implemented a supervised-only baseline with the same model capacity: A sentence encoder followed by a logistic regression model. Weight decay with rate 1e-3 is additionally applied to avoid serious over-fitting in data scarce scenarios. As can be seen from Table~\ref{tab:cpc-SSL-LargeMovieReview}, our semi-supervised learning approach significantly outperforms its supervised baseline, validating the effectiveness of our semi-supervised models. 

\section{Related Works}
\label{sec:related works}
The significant practical importance of semi-supervised learning in modern data analysis has motivated a large body of research. The developed techniques are now roughly categorized into the following types.

\subsection{Graph-Based Methods}
These methods require a graph describing similarity between data points, each data a node. Similarity can be based on Euclidean distance, Mahalanobis distance or task-specific metrics~\cite{weston12ssl}. \cite{zhu02lp} propose label propagation (LP). LP iteratively propagates the class posterior of a node to its neighbours until the process finally reach an equilibrium. Though proved to be successful on some benchmarks, these methods requires a pre-constructed graph and the their performance is largely affected by the choice of similarity. High dimensionality of raw data in modern applications also poses great challenges to these approaches. \cite{haeusser17ssl} tried to address these problems by introducing a deep feature extractor and build the graph in the latent space. Associations between labeled and unlabeled data is modeled through the probability that a two-step random walk would start and end at labeled samples of the same class, via one intermediate unlabeled point. \cite{konstantinos18ssl} dynamically create a graph over latent representations of labeled and unlabeled samples of a training batch, and operates LP to estimate high and low density regions. Despite the endeavour made, calculating similarity between each input data points and operates on the graph make these kind of approaches hard to scale to modern large scale applications. The computational complexity of these methods goes high up to $\mathcal{O}(N^3)$, where $N$ is the number of training data.


\subsection{Perturbation-Based Approaches}
Ability to resist noise applied to inputs can improve generalization of a classifier. \cite{bachman14pseudo-ensumble} first regularize the input-output mapping to be consistent under noise corruption in semi-supervised learning. \cite{rasmus15ladder} proposed $\Gamma$ model, in which each data is evaluated with and without noise, deviation between original data prediction and noise corrupted data prediction is punished. This encourages local consistency, pushing decision boundaries away from high density areas. Type of noised added has also been extensively explored, \cite{bachman14pseudo-ensumble} used different dropout masks. Random Guassian noise is applied in~\cite{tarvainen17meanteacher}. Recently, adding adversarial perturbation~\cite{miyatoMKI17adv-semi} gives impressive result in SSL. Perturbation based approaches do not change the model architecture of original classifier, Gold-standard classification model can thus be easily exploited. Besides, No additional computational overhead is introduced by adding the consistency regularization. These models are computational friendly and thus can be good candidates for real word application. However, In these approaches the model behave both as a teacher and a student. Prediction of the teacher is used as target for the student. The generated targets may well be incorrect, thus these approaches suffer from confirmation bias. Although there have been researches on alleviating the problem. How destructive confirmation bias will be for a specific application and how to avoid the problem is still not clear. Perturbation based methods are compatible to our methods, their combination may produce even better outcomes. The analysis of their combined effects will be left for future work.

Closely related is self-supervision, one of the earliest ideas for semi-supervised learning. In~\cite{scudder65self-sup}, a classifier trained with only labeled data is used to predict the unlabeled samples. Unlabeled examples that are confidently predicted, as well as its predicted label will be add into the training set for next round training.  \cite{lee13pseudolabel} brought the idea of self-supervision into deep neural network training. Generally these models enjoy good performance. However, self-supervision methods heavily depends on performance of classifier trained on the initial labeled data. Classification accuracy for a specific task is not assured. Dynamically adding training samples and periodically retraining also pose great challenge for large scale applications. 

\subsection{Generative Models}
Generative models fits the joint distribution $p(x, y) = p(y|x) p(x)$ instead of just model a conditional $p(y)$. Semi-supervised learning problem is then recognised as a specialised missing data imputation task. Early examples applying generative models for semi-supervised learning includes mixture of Gaussian and hidden Markov models. These models are restricted to shallow structure of hidden variables and are not flexible enough for complex data. \cite{kingma14dgm} first introduce deep generative model variational auto-encoder into SSL and get good results on simple visual data benchmarks. To make the approximated posterior more flexible, ~\cite{maaloe16adgm} added axillary variable on the latent factors, take expectation on which results in much flexible approximated posterior. Variant follows~\cite{li15mmdgm}. Auto-regressive generative models are also utilized in SSL, ~\cite{dai15semi-autoregressive} explored possibility of modelling the data distribution $p(x)$ with a sequence autoencoder to help classification. Another kind of generative model that widely appliedin SSL is generative adversarial networks(GAN)~\cite{goodfellow14gan}. \cite{salimense16improvedgan} encouraged the discriminator to not only distinguish real from fake inputs, but also identify the class of real samples. Later on, Conditional GANs were used to generate data pairs $(x, y)$, which can also be used for SSL training.\cite{li2017triplegan}. GAN based SSL methods got promising results on several benchmarks yet the challenges of adversarial optimization leave space for improving. Utilizing generative model to capture the structure of unlabeled data is generally promising, however, preserving variation of the inputs and pixel-wisely reconstruct or generate data requires models of unnecessarily large representational power and complexity, hinders their application on large scale problems. Besides, it has already been shown that the very details or local features are not relevant to down streaming tasks like classification. Using models that captures longer range structure of data might be better for SSL tasks.

\section{Conclusion and Future Works}

We have developed new models for semi-supervised learning that allow us to improve the quality of prediction by exploiting contrastive predictive coding, which capture long range or coarse-grained data information and avoid wasting too much model capacity on the very details. We have developed efficient algorithm for approximate Bayesian inference in these models and demonstrated that they are amongst the most competitive models currently available for semi-supervised learning in large scale applications. We hope that these results stimulate the development of even more powerful semi-supervised learning methods.

For classification tasks, one area of interest is to combine such methods with max-margin principles, which has been show to be effective discriminating samples form different classes. Since our model contains an classification component, we can readily combine Bayesian SVMs, which forms a promising avenue for future exploration. Beside, the proposed methods are compatible to other SSL approaches. Developing model that utilize the salient factors extracted by contrastive predictive coding, as well resist to perturbations in a sense may also be fruitful. 
\newpage

\bibliographystyle{named}
\bibliography{ijcai19}

\end{document}